\documentclass[10pt,twocolumn,letterpaper]{article}

\usepackage[pagenumbers]{cvpr} 

\usepackage[dvipsnames]{xcolor}

\definecolor{cvprblue}{rgb}{0.21,0.49,0.74} 
\colorlet{dark-green}{green!70!black} 

\usepackage[pagebackref,breaklinks,colorlinks,allcolors=cvprblue]{hyperref} 

\usepackage{graphicx}
\usepackage{amsmath}
\usepackage{amssymb}
\usepackage{multirow} 
\usepackage{booktabs} 
\usepackage{wasysym}	
\usepackage{microtype} 
\usepackage{bm}	 
\usepackage{float}	 

\def\paperID{1990}	 
\def\confName{CVPR}	
\def\confYear{2025}	

\title{Expanding Zero-Shot Object Counting with Rich Prompts}
\author{
Huilin Zhu$^{1,2}$, Senyao Li$^{1}$, Jingling Yuan$^{1,\ast}$, Zhengwei Yang$^{3}$, Yu Guo$^{4,2}$, Wenxuan Liu$^{5,1}$, \\ Xian Zhong$^{1,\ast}$, and Shengfeng He$^{2}$ \\
$^{1}$ Hubei Key Laboratory of Transportation Internet of Things, 
Wuhan University of Technology \\
$^{2}$ School of Computing and Information Systems, Singapore Management University \\
$^{3}$ School of Computer Science, Wuhan University \\
$^{4}$ School of Navigation, 
Wuhan University of Technology \\
$^{5}$ School of Computer Science, Peking University \\
{\tt\small yjl@whut.edu.cn, zhongx@whut.edu.cn}
}

\begin{document}

\maketitle

\begin{abstract}
Expanding pre-trained zero-shot counting models to handle unseen categories requires more than simply adding new prompts, as this approach does not achieve the necessary alignment between text and visual features for accurate counting. We introduce RichCount, the first framework to address these limitations, employing a two-stage training strategy that enhances text encoding and strengthens the model’s association with objects in images. 
RichCount improves zero-shot counting for unseen categories through two key objectives: (1) enriching text features with a feed-forward network and adapter trained on text-image similarity, thereby creating robust, aligned representations; and (2) applying this refined encoder to counting tasks, enabling effective generalization across diverse prompts and complex images. 
In this manner, RichCount goes beyond simple prompt expansion to establish meaningful feature alignment that supports accurate counting across novel categories. Extensive experiments on three benchmark datasets demonstrate the effectiveness of RichCount, achieving state-of-the-art performance in zero-shot counting and significantly enhancing generalization to unseen categories in open-world scenarios.

\end{abstract}

\section{Introduction}

Object counting is a fundamental task in computer vision with applications ranging from crowd, vehicle, and cell counting~\cite{tyagi2023degpr, arteta2016counting, mundhenk2016large, babu2022completely}. Traditional methods for counting rely on models trained for specific object categories, limiting their ability to generalize to new or unseen categories. Class-agnostic counting addresses this limitation by training models on known categories that can generalize to a wider range of unseen objects.
Few-shot learning~\cite{liu2022countr,djukic2023low,lu2019class,ranjan2021learning,WangX0024} has emerged as a leading approach for class-agnostic object counting. It leverages a small number of annotated bounding boxes to model the relation between the boxes and the image, enabling the identification of unseen objects. These models exploit the strong correlation between visual prompts and object representations, demonstrating effective performance across unseen categories.

\begin{figure}
	\centering
	\includegraphics[width = \linewidth]{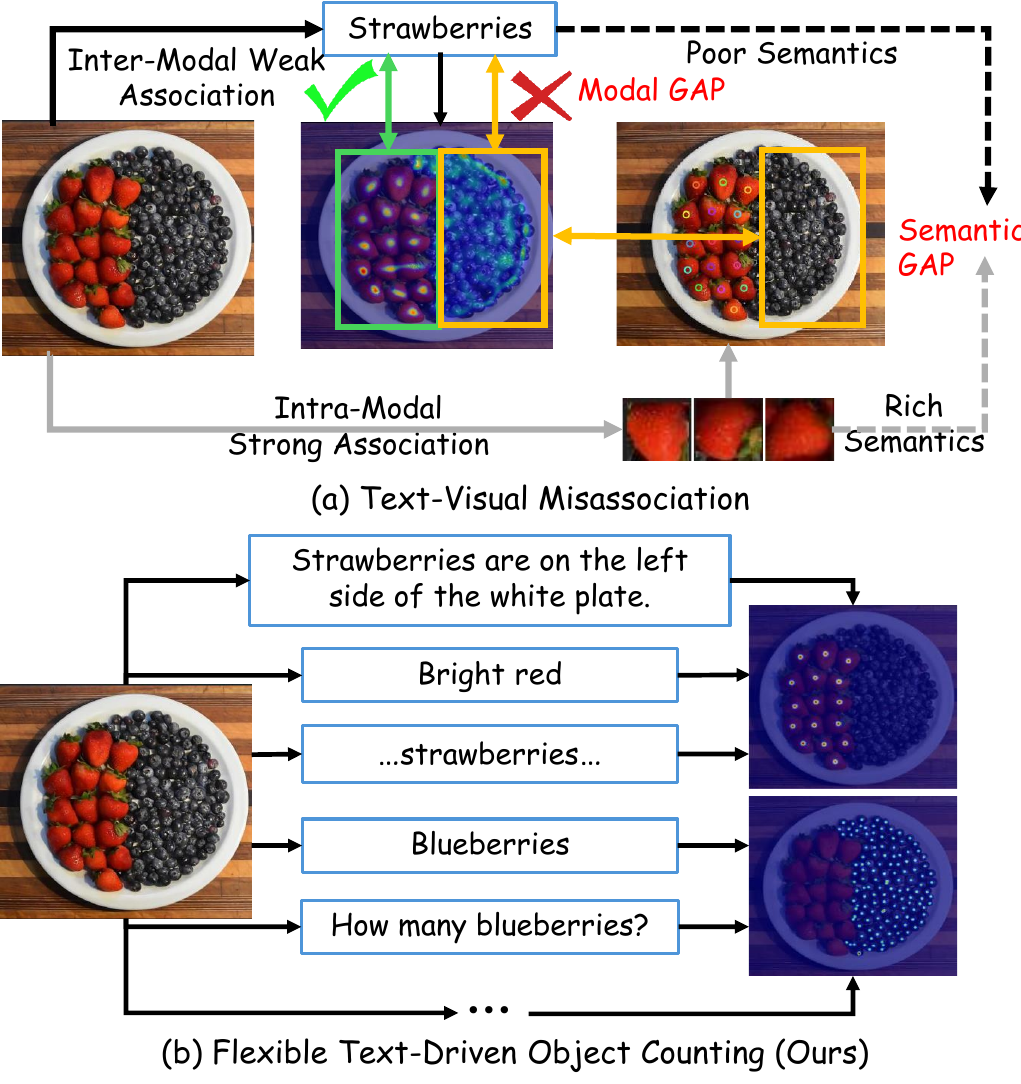}
	\caption{\textbf{Illustration of Text-Visual Association.} (a) Text-based counting methods (ClipCount~\cite{jiang2023clip}) often result in non-specific category estimations, whereas visual prompts (T-Rex~\cite{jiang2023t}) mitigate this issue. A natural modality and semantic gap exists between text and visual prompts. (b) Our method addresses this misalignment, enabling the use of diverse text prompts as inputs.}
	\vspace{-5pt}
	\label{fig:1}
\end{figure}

However, in practical scenarios, images of unknown categories often lack annotated bounding boxes, making text-based methods, known as zero-shot counting, a promising solution. Techniques like CLIP-Count~\cite{jiang2023clip} and VLCount~\cite{kang2023vlcounter} generate density maps by modeling pixel-level relations between text and images. Yet, they face a significant challenge due to the modal gap arising from the disparity between textual and visual representations. Unlike visual prompts that naturally align with image features, text prompts introduce misalignment complicating accurate counting. 
Methods like CounTX~\cite{amini2023open} attempt to address this issue by improving text descriptions, but they do not fully mitigate the underlying modal gap. Recent approaches, such as ZSC~\cite{xu2023zero} and VA-Count~\cite{zhu2024zero}, incorporate visual exemplars within images to bridge the textual and visual features, reducing the modal gap. However, visual prompts, while reducing the gap, often introduce noise, which negatively impacts model performance. As a result, zero-shot methods tend to underperform compared to few-shot methods, which rely on precise visual prompts. The key difference between these approaches lies in their use of textual prompts for zero-shot counting versus precise visual prompts for few-shot counting.

\cref{fig:1}(a) illustrates that text prompts may occasionally identify objects outside the intended category, a phenomenon less common with bounding box prompts. This discrepancy arises because visual prompts, such as bounding boxes, directly target objects within the image~\cite{WangX0024}, while text prompts capture semantically similar objects, highlighting the modal gap between textual and visual features.

The difference between text and visual prompts extends beyond modal disparity. Bounding box-based visual prompts convey richer semantic information, including attributes such as color, shape, and other appearance details, which remain consistent with the overall image style. In contrast, text prompts typically provide only categorical information. In open-world counting scenarios, where unseen categories are encountered, relying solely on category-level text is insufficient for accurately identifying objects, making text prompts less effective. Additionally, zero-shot methods are generally limited to category-level text during inference, restricting the model's ability to leverage more complex and flexible text prompts for improved counting performance.

From this analysis, we identify three primary challenges in zero-shot counting: 
1) modal gap between text prompts and image features, 
2) limited semantic richness of category-level text, and 
3) rigidity of text inputs during inference. Addressing these challenges requires not just generating generic text prompts but developing prompts that are closely aligned with image features.

To address these issues, we propose RichCount, a two-stage training strategy with two main objectives: enhancing text encoding and improving the model's ability to associate prompts with objects in the image. In the first stage, RichCount generates enriched text features by training a feed-forward network based on the similarity between text and image features. This is followed by training an adapter to refine the encoding process, producing more aligned and robust feature representations. The second stage uses this enhanced encoder to train the model for counting tasks, enabling it to associate prompts with objects effectively and recognize unseen categories using flexible prompts. Through this structured approach, RichCount ensures that enriched text representations align closely with visual features, advancing zero-shot counting from basic prompt generation to adaptive, task-specific object counting in diverse scenarios.


In summary, our contributions are threefold:

\begin{itemize}
	\item We investigate the relation between text prompts and zero-shot counting, identifying key challenges such as modal disparity, limited semantic richness, and prompt flexibility. Our findings offer insights for both zero-shot counting and other visual-text understanding tasks.

	\item We propose RichCount, a novel two-stage training strategy that enhances text representations, aligns visual and textual features, and enables robust prompt processing for zero-shot counting.

	\item Extensive experiments across three object counting datasets validate the effectiveness and scalability of RichCount, demonstrating its state-of-the-art performance in open-world object counting tasks.

\end{itemize}

\section{Related Work}

\paragraph{Few-shot Object Counting} 
Few-shot Object Counting has made significant strides in addressing the challenge of limited annotated data. CounTR~\cite{liu2022countr} employs transformers for scalable and efficient counting, while LOCA~\cite{djukic2023low} improves generalization by enhancing feature representation and adapting exemplars. Earlier methods, such as GMN~\cite{lu2019class}, framed class-agnostic counting as a matching problem, a concept further refined by BMNet~\cite{shi2022represent} using bilinear matching for more precise similarity assessments. FamNet~\cite{ranjan2021learning} incorporated ROI Pooling to improve feature extraction, and CACViT~\cite{WangX0024} integrated Vision Transformers (ViT) into object counting architectures, resulting in additional performance improvements. CountGD~\cite{amini2024CountGD} builds upon the powerful vision-language model GroundingDINO~\cite{Liu2023DINO} to enhance the generality and accuracy of open-vocabulary object counting in images.

\vspace{-10pt}
\paragraph{Zero-shot Object Counting} 
Zero-shot Object Counting~\cite{xu2023zero,xu2023zerol}, which utilizes text prompts instead of visual exemplars, offers flexible object specification without needing training data in target categories. Approaches like CLIP-Count~\cite{jiang2023clip} leverage CLIP to separately encode text and images for semantic alignment, while VLCount~\cite{kang2023vlcounter} enhances text-image alignment. PseCo~\cite{huang2023point} introduces a SAM-based framework for segmentation, dot mapping, and detection, expanding applicability but with high computational demands. Despite the potential of these methods, they often face alignment challenges between visual and textual information, which impacts accuracy. This paper addresses these limitations by improving the alignment between visual and textual prompts, leading to more precise zero-shot object counting.

\vspace{-10pt}
\paragraph{Multi-modal Large Language Models} 
Multi-modal Large Language Models (MLLMs) have driven major advancements in several fields. Systems such as Kosmos-2~\cite{peng2023kosmos}, Shikra~\cite{chen2023shikra}, GPT4RoI~\cite{zhang2023gpt4roi}, and VisionLLM~\cite{wang2024visionllm} combine generative Large Language Models (LLMs) with localization tasks, enabling region-level human-model interactions. Building on these foundations, recent models like LISA~\cite{lai2024lisa}, GLaMM~\cite{rasheed2024glamm}, and PixelLM~\cite{ren2024pixellm} introduce pixel-level segmentation, further pushing the boundaries of multi-modal capabilities. Despite these advances, the application of MLLMs~\cite{TheC3, achiam2023gpt} in specialized domains, such as image quality assessment and visual grounding, remains underexplored.

However, many existing methods still rely on annotated bounding boxes, which limits their applicability in real-world scenarios where such annotations are costly or unavailable. This dependence reduces the flexibility of models, particularly for unseen categories, highlighting the need for more adaptable solutions.

\begin{figure*}
	\centering
	\includegraphics[width = \linewidth]{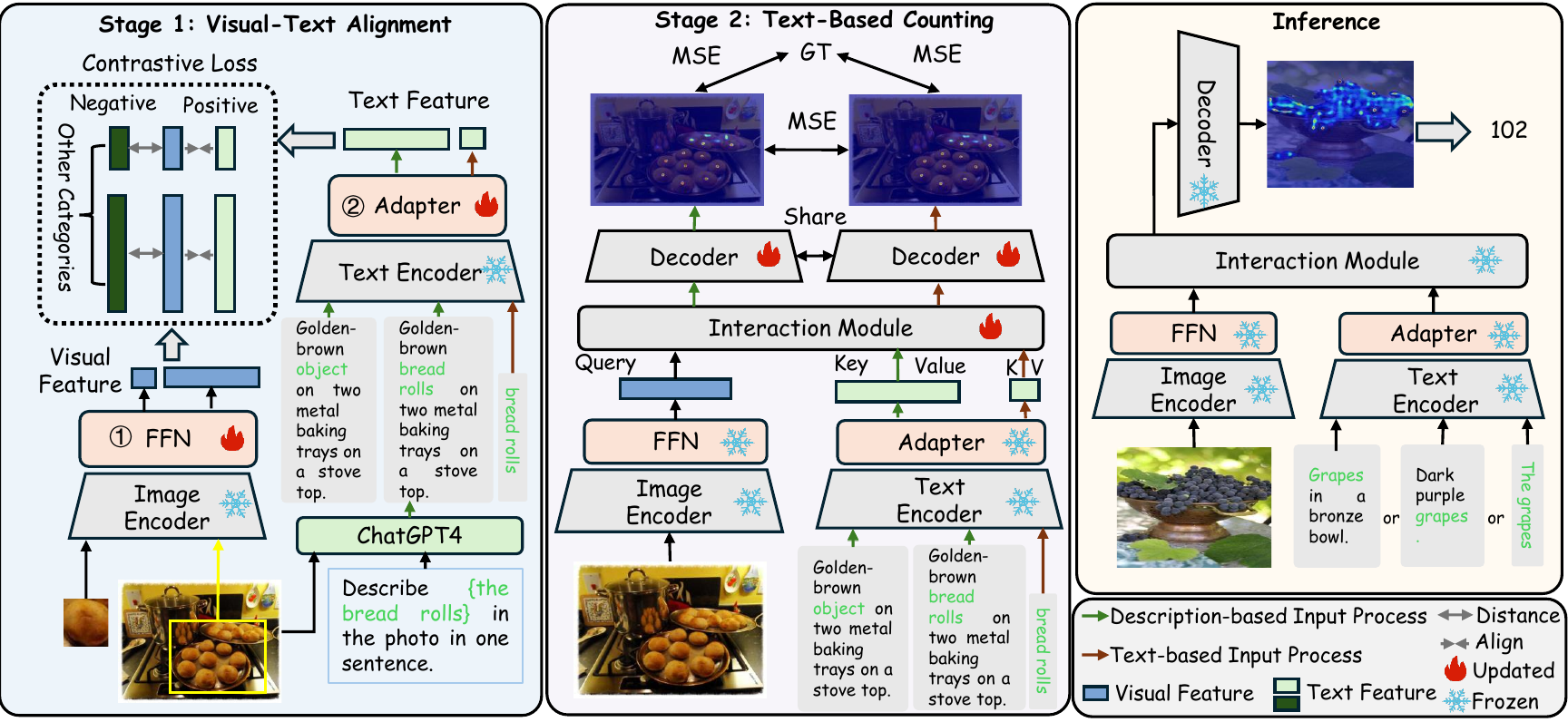}
	\caption{\textbf{Overview of the Proposed Method.} The framework consists of two training stages: (1) {Visual-Text Alignment}, which utilizes ChatGPT to generate descriptive text for image categories. To align features, an FFN is added to the CLIP visual encoder, and an adapter is integrated into the CLIP text encoder; (2) {Text-Based Counting}, which freezes the encoders and trains the interaction module and decoder to ensure consistency between density maps generated from text descriptions and their corresponding textual inputs. During inference, the model generates density maps based on diverse textual prompts.}
	\label{fig:2}
	\vspace{-5pt}
\end{figure*}

\section{Proposed Method}

Zero-shot object counting is designed to estimate the number of objects specified by a textual prompt, with the distinct condition that the categories in the training ($X_\mathrm{train}$), validation ($X_\mathrm{val}$), and testing ($X_\mathrm{test}$) sets do not overlap, \textit{i.e.}, $X_\mathrm{train} \cap X_\mathrm{val} \cap X_\mathrm{test} = \emptyset$. To overcome challenges such as modal gaps, limited semantic depth in text prompts, and inflexible textual inputs, as shown in \cref{fig:2}, we propose a two-stage framework comprising \textbf{Visual-Text Alignment} (\cref{Sec.3.2}) and \textbf{Text-Based Counting} (\cref{Sec.3.3}).

In the Visual-Text Alignment stage, text representations are first enriched using an MLLM, providing semantic-rich descriptions $T_d$ that surpass simple category labels $T_p$. These enriched text features are then aligned with image features using enhanced encoders, $f_v(\cdot)$ for images and $f_t(\cdot)$ for text. This alignment is achieved through contrastive learning, guided by the objective $O_1$:
\begin{equation} 
	\label{eq:o1}
	\mathrm{O_1} = 
	\begin{cases}
	\max \mathrm{sim} \left(f_v(V), f_t(T) \right), \\
	\min \mathrm{sim} \left(f_v(V), f_t(T_n) \right),
	\end{cases}
\end{equation}
where $V$ denotes the input image, $T$ represents textual prompts ($T_p$ or $T_d$), $T_n$ corresponds to negative samples from other categories, and $\mathrm{sim}(\cdot)$ quantifies feature similarity.

In the Text-Based Counting stage, a counter generates density maps from text inputs. Given an image $I$, prompts $T_p$ and $T_d$ are processed by $M_\mathrm{fuse}(\cdot)$ and $f_d(\cdot)$ to generate density maps $D_t$ and $D_d$.

\begin{equation}
	\label{eq:d1}
	D = f_d \left(M_\mathrm{fuse} \left(f_v(V), f_t(T) \right) \right).
\end{equation}
The objective $O_2$ minimizes the discrepancy between these predicted maps and the ground truth $D_g$, while ensuring consistency between $D_t$ and $D_d$:

\begin{equation} 
	\label{eq:o2}
	\mathrm{O_2} = \min \mathrm{Diff} \left(D_t, D_g, D_d \right),
\end{equation}
where $\mathrm{Diff}(\cdot)$ quantifies the discrepancy.

\subsection{Visual-Text Alignment}
\label{Sec.3.2}
\paragraph{Description Augmentation.}
Given an image $I$ and a category name $T_p$, an MLLM $G(\cdot)$ generates a detailed description $T_d$ by processing the image $I$ and a prompt $P_t$ containing the category $T_p$:
\begin{equation}
	T_d = G \left(I, P_t \right).
\end{equation}
This enriched description captures not only the category information but also attributes such as appearance and location, thereby enhancing the semantic richness beyond simple category names.

\begin{figure}
	\centering
	\includegraphics[width = \linewidth]{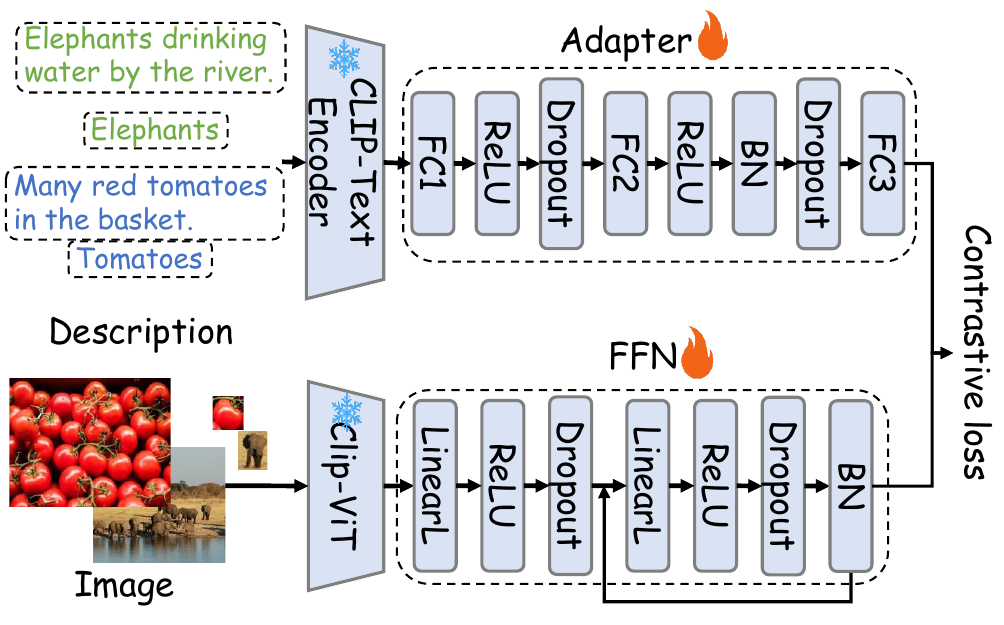}
	\caption{\textbf{Illustration of the Alignment Strategy.} An adapter refines text embeddings, and an FFN processes visual features aligned via contrastive loss for cross-modal understanding.}
	\label{fig:3}
	\vspace{-5pt}
\end{figure}

\vspace{-10pt}
\paragraph{Alignment.}
With the enriched descriptions generated, the alignment process refines the correspondence between image and text features. As illustrated in \cref{fig:3}, this strategy builds on the foundational CLIP architecture, incorporating a feed-forward network (FFN) and an adapter to optimize text-image feature alignment. The inputs consist of visual elements $V = \{ V_p, V_r \}$, representing visual prompts and cropped regions, and textual elements $T = \{ T_p, T_d, T_d' \}$, where $T_p$ denotes the category name, $T_d$ represents the enriched description, and $T_d'$ replaces the category name with ``object''.

The contrastive loss function used to train the FFN and the adapter is defined as:

\begin{equation}
\label{eq:e}
	\Delta_p = \| e_{\mathrm{img}} - e_p \|, \quad
	\Delta_n = \| e_{\mathrm{img}} - e_n \|, 
\end{equation} 
\begin{equation} 
\resizebox{0.9\columnwidth}{!}{$
	\mathcal{L}_c = \frac{1}{2N} \sum_{i=1}^{N} \big[ y^i (\Delta_p^i)^2 + (1 - y^i) (\max(0, m - \Delta_n^i))^2 \big], $}
\label{eq:contrastive_loss}
\vspace{1mm}
\end{equation} 
where $y^i \in \{0, 1\}$ indicates the match status of pairs, $m$ is the margin for separation, and $N$ is the total number of pairs.

\vspace{-10pt}
\paragraph{Feature Enhancement via FFN.}
Initially, an FFN is integrated into the CLIP visual encoder $C_v(\cdot)$:
\begin{equation} 
	f_v(\cdot) = \mathrm{FFN}(C_v(\cdot)),
\end{equation}
During this phase, $C_v$ is frozen to preserve pre-trained features. The embeddings are:
\begin{equation}
	e_{\mathrm{img}} = f_v(V), \quad e_p = C_t(T), \quad e_n = C_t(T_n).
\end{equation}
These embeddings are used to train the FFN, as delineated in \cref{eq:e} and \cref{eq:contrastive_loss}.

\vspace{-10pt}
\paragraph{Adapter Training for Textual Feature Refinement}
Subsequently, an adapter is integrated into the text encoder after enhancing the visual features:
\begin{equation} 
	f_t(\cdot) = \mathrm{Adapter}(C_t(\cdot)).
\end{equation}
During this phase, both the FFN and CLIP encoder are frozen, focusing training solely on the adapter to synchronize with updated visual outputs:
\begin{equation} 
	e_p = f_t(T), \quad e_n = f_t(T_n).
\end{equation}
The adapter is trained through the alignment of visual and text features by applying the contrastive loss defined in \cref{eq:e} and \cref{eq:contrastive_loss}.

\begin{table*}
	\centering
	\footnotesize
	\setlength{\tabcolsep}{8pt}
	\begin{tabular}{clcccccccc}
	\toprule[1.1pt]
	\multirow{2}[2]{*}{Scheme} & \multirow{2}[2]{*}{Method} & \multirow{2}[2]{*}{Venue} & \multirow{2}[2]{*}{Exemplar} & \multicolumn{2}{c}{Val Set} & \multicolumn{2}{c}{Test Set} & \multicolumn{2}{c}{Avg} \\
	\cmidrule(lr){5-6} \cmidrule(lr){7-8} \cmidrule(lr){9-10}
	& & & & MAE & RMSE & MAE & RMSE & MAE & RMSE \\
	\midrule
	\multirow{4}{*}{Reference-less}
	& FamNet~\cite{ranjan2021learning} & CVPR'21 & None & 32.15 & 98.75 & 32.27 & 131.46 & 32.21 & 115.11 \\
	& RCC~\cite{hobley2022learning} & CVPR'22 & None & \underline{17.49} & \underline{58.81} & 17.12 & \underline{104.53} & 17.31 & \underline{81.67} \\
	& CounTR~\cite{liu2022countr} & BMVC'23 & None & 18.07 & 71.84 & \textbf{14.71} & 106.87 & \textbf{16.39} & 89.36 \\
	& LOCA~\cite{djukic2023low} & ICCV'23 & None & \textbf{17.43} & \textbf{54.96} & \underline{16.22} & \textbf{103.96} & \underline{16.83} & \textbf{79.46} \\
	\midrule
	\multirow{7}{*}{Few-shot}
	& FamNet~\cite{ranjan2021learning} & CVPR'21 & Visual Exemplars & 24.32 & 70.94 & 22.56 & 101.54 & 23.44 & 86.24 \\
	& CFOCNet~\cite{yang2021class} & WACV'22 & Visual Exemplars & 21.19 & 61.41 & 22.10 & 112.71 & 21.65 & 87.06 \\
	& CounTR~\cite{liu2022countr} & BMVC'22 & Visual Exemplars & 13.13 & 49.83 & 11.95 & 91.23 & 12.54 & 70.53 \\
	& PseCo~\cite{huang2023point} & CVPR'23 & Visual Exemplars & 15.31 & 68.34 & 13.05 & 112.86 & 14.18 & 90.60 \\
	& LOCA~\cite{djukic2023low} & ICCV'23 & Visual Exemplars & \underline{10.24} & \underline{32.56} & 10.97 & \underline{56.97} & \underline{10.61} & \underline{44.77} \\
	& CACViT~\cite{WangX0024} & AAAI'24 & Visual Exemplars & 9.13 & 10.63 & 48.96 & 37.95 & 10.94 & 52.99 \\
	& CountGD~\cite{amini2024CountGD} & NeurIPS'24 & Visual \& Text & \textbf{7.10} & \textbf{26.08} & \textbf{5.74} & \textbf{24.09} & \textbf{6.42} & \textbf{16.25}\\
	\midrule
	\multirow{8}{*}{Zero-shot} 
	& ZSC~\cite{xu2023zero} & CVPR'23 & Text & 26.93 & 88.63 & 22.09 & 115.17 & 24.51 & 101.90 \\
	& VA-Count~\cite{zhu2024zero} & ECCV'24 & Text & 17.87 & 73.22 & 17.88 & 129.31 & 17.87 & 101.26 \\
	& VLCount~\cite{kang2023vlcounter} & AAAI'24 & Text & 18.06 & 65.13 & 17.05 & 106.16 & 17.56 & 85.65 \\
	& CounTX~\cite{amini2023open} $\dag$ & BMVC'23 & Text & \textbf{16.99} & 61.67 & 17.29 & 112.50 & \underline{17.15} & 87.09 \\
	& CLIP-Count~\cite{jiang2023clip} $\dag$ & ACM MM'23 & Text & 19.85 & 67.69 & 17.19 & 103.44 & 18.52 & 85.57 \\
	& CLIP-Count~\cite{jiang2023clip} $\dag$ & ACM MM'23 & Description & 19.52 & 67.80 & 17.49 & 104.60 & 18.51 & 86.20 \\
	& RichCount (Ours) & & Text & 18.65 & \underline{58.55} & \underline{16.37} & \underline{102.48} & 17.51 & \underline{80.51}\\
	& RichCount (Ours) & & Description & \underline{17.68} & \textbf{57.24} & \textbf{15.78} & \textbf{99.65} & \textbf{16.73} & \textbf{78.45}\\
	\bottomrule[1.1pt]
	\end{tabular}
	\caption{\textbf{Quantitative Results on \textsc{FSC-147}.} Methods are compared using text, visual, and hybrid prompts, with {Avg} denoting the average performance across test and validation sets. Models reproduced in this study are marked with $\dag$, and the best and second-best results are highlighted in {bold} and {underlined}, respectively.} 
	\label{tab:ExpSOTA}
	\vspace{-5pt}
\end{table*}

\subsection{Text-Based Counting}
\label{Sec.3.3}
Building on the aligned encoders from the previous stage, the second stage freezes the visual and text encoders, $f_v(\cdot)$ and $f_t(\cdot)$, to focus on training the Interaction Module $M_\mathrm{fuse}(\cdot)$ and decoder $f_d(\cdot)$. This stage models the interactions between textual prompts and target objects in images. Text inputs, category name $T_p$, detailed descriptions $T_d$, and generalized descriptions $T_d'$ where "object" replaces the specific category, are paired with the original image $I$. This pairing aims to enhance the model’s ability to generalize across different textual representations and improve accuracy in interpreting diverse textual contexts.


\vspace{-10pt}
\paragraph{Feature Fusion.}
The Interaction Module fuses multi-modal information by treating image embeddings $\bm{e}_\mathrm{img} = f_v(I)$ as queries and text embeddings $\bm{e}_\mathrm{txt} = f_t(T)$ as keys and values. The fused features are computed as:
\begin{equation} 
	\bm{e}_\mathrm{fuse} = M_\mathrm{fuse} \left(\bm{e}_\mathrm{img}, \bm{W}^k \bm{e}_\mathrm{txt}, \bm{W}^v \bm{e}_\mathrm{txt} \right),
\end{equation}
where $\bm{W}^k$ and $\bm{W}^v$ are learnable projection weights for keys and values, ensuring alignment and dimensional consistency. This fusion bridges the gap between text and image features, enabling robust counting in zero-shot settings.

\vspace{-10pt}
\paragraph{Density Map Generation.}
The decoder generates a density map from the fused features:
\begin{equation} 
	D = f_d \left(\bm{e}_\mathrm{fuse} \right).
\end{equation}
The density loss, denoted as $\mathcal{L}_D$, is computed as the mean squared error (MSE) between two density maps, $D^a$ and the ground truth density map $D^b$:
\begin{equation} 
\label{eq:ld}
	\mathcal{L}_D \left(D^a, D^b \right) = \frac{1}{HW} \sum_{i=1}^{H} \sum_{j=1}^W \left( D^a_{i,j} - D^b_{i,j} \right)^2,
\end{equation}
where $H$ and $W$ are the height and width of the image.

\vspace{-10pt}
\paragraph{Total Loss.}
To ensure consistent predictions across various textual inputs, the total loss $\mathcal{L}_t$ is defined as:
\begin{equation} 
\label{eq:lt}
	\mathcal{L}_t = \sum_{a} \mathcal{L}_D \left(D^a, D^g \right) + \sum_{(a,b)} \mathcal{L}_D \left(D^a, D^b \right),
\end{equation}
where $a \in \{t, d, d'\}$ and $(a,b) \subset \{t, d, d'\}$. $D^t$, $D^d$, and $D^{d'}$ correspond to density maps generated from category labels, detailed descriptions, and generalized descriptions, respectively. The index $x$ iterates over category labels $t$, detailed descriptions $d$, and generalized descriptions $d'$, aligning each with the ground truth density map $D^g$. The unordered pairs $(x,y)$ include $(t, d)$, $(t, d')$, and $(d, d')$ to ensure consistency between different textual descriptions.



\subsection{Inference}
\label{Sec.3.4}

During inference, the model processes input images $I$ along with various textual inputs $T_\mathrm{in}$, including category names, detailed descriptions, or attribute-based prompts, enabling zero-shot object counting for unseen categories. The density map is calculated as:
\begin{equation} 
	D_\mathrm{out} = f_d \left(M \left(f_v(I), f_t(T_\mathrm{in}) \right) \right),
\end{equation}
and the total object count is obtained by summing the pixel values in the density map:
\begin{equation} 
	\mathrm{Count} = \sum_{i = 1}^{H} \sum_{j = 1}^W D_\mathrm{out}(i, j),
\end{equation}
where $H$ and $W$ represent the map dimensions.

\section{Experimental Result}

\subsection{Datasets and Implementation Details}

\paragraph{Datasets.}
\textsc{FSC-147}~\cite{hobley2022learning} dataset is a class-agnostic counting dataset comprising 6,135 images across 147 classes, designed specifically for zero-shot counting. The dataset features non-overlapping subsets for training, validation, and testing, with dot annotations provided for precise object localization. 
Descriptions are extended from class names and images, with class text replaced or enriched to improve generalization and textual input robustness.

\textsc{CARPK}~\cite{hsieh2017drone} dataset contains 89,777 car instances in 1,448 parking lot images, making it an ideal benchmark for evaluating cross-dataset transferability.

\textsc{ShanghaiTech}~\cite{zhang2016single} dataset is a crowd counting dataset with two parts: Part A (\textsc{SHA}) consisting of 482 images and Part B (\textsc{SHB}) consisting of 716 images. Each part includes 400 training images, though cross-part evaluations are challenging due to differences in data collection methods.

\vspace{-10pt}
\paragraph{Implementation Details.} 
In all experiments, we used a fixed image encoder and a text encoder initialized with pre-trained CLIP (ViT-B/16). Following ClipViT, we introduced a context-aware FFN with an input dimension of 512, structured as a fully connected network featuring hidden layers, batch normalization, and ReLU activations. The CLIP Text Transformer processes text prompts up to 77 tokens, each embedded in a 512-dimensional space. In the contrastive learning for image-text alignment, the margin is set to 1. We trained all datasets for 200 epochs with a batch size of 64 on an NVIDIA RTX L40 GPU.

\subsection{Comparison with State-of-the-Art Methods}

\paragraph{Quantitative Results on \textsc{FSC-147}.}
RichCount was evaluated on \textsc{FSC-147} and compared to state-of-the-art methods, as shown in \cref{tab:ExpSOTA}. In the zero-shot setting, RichCount achieved the best and second-best performance, with an MAE of 15.78 on the test set, significantly outperforming other models. Compared to ClipCount~\cite{jiang2023clip}, RichCount reduced the test set MAE by 1.71 and outperformed its own variant trained with class labels as text input, achieving a 0.59 MAE improvement.
RichCount also demonstrated superior generalization on the unseen-class test set, maintaining an RMSE below 100, reflecting its ability to overcome cross-modal alignment challenges and enhance semantic representation. In contrast, ClipCount struggled with more complex textual descriptions, increasing its test set MAE by 0.3 when using RichCount’s descriptions, highlighting its encoder's limitations in handling enriched text and providing complementary information for visual samples.
While CounTX~\cite{amini2023open} achieved the best MAE on the validation set, its performance on unseen categories was less competitive due to reliance on simple class labels, which lack semantic richness. Despite a performance gap compared to few-shot methods, RichCount surpassed reference-less approaches, demonstrating the effectiveness of enriched text and image-text alignment in improving counting accuracy.

\begin{table}
	\centering
	\footnotesize
	\setlength{\tabcolsep}{5pt}
	\begin{tabular}{lccccc}
	\toprule[1.1pt]
	\multirow{2}[2]{*}{Method} & \multirow{2}[2]{*}{Venue} &\multirow{2}[2]{*}{Exemplar} & \multicolumn{2}{c}{FSC $\to$ \textsc{CARPK}} \\ 
	\cmidrule(lr){4-5} 
	& & & MAE & RMSE \\
	\midrule 	
	FamNet~\cite{ranjan2021learning}& CVPR’21 & Visual & 28.84 & 44.47 \\
	BMNet~\cite{shi2022represent}& CVPR’22 & Visual & 14.41 & 24.60 \\
	BMNet+~\cite{shi2022represent} & CVPR’22& Visual & 10.44 & 13.77 \\
 	\midrule 	
	RCC~\cite{hobley2022learning} & CVPR’22 & Text & 21.38 & 26.61 \\
	CLIP-Count~\cite{jiang2023clip} & ACM MM’23 & Text & 13.59 & 18.30 \\
	RichCount (Ours) & & Description & \textbf{9.91} & \textbf{13.28} \\	
	\bottomrule[1.1pt]
	\end{tabular}
	\caption{\textbf{Comparison of Our Method with State-of-the-Art Zero-Shot and Few-Shot Approaches on \textsc{CARPK}.}}
	\label{tab2}
	\vspace{-5pt}
\end{table}

\vspace{-10pt}
\paragraph{Quantitative Results on \textsc{CARPK}.}
To assess the cross-dataset generalization of our model, we tested it on \textsc{CARPK}. The model was trained on \textsc{FSC-147} and evaluated on \textsc{CARPK} without fine-tuning. As shown in \cref{tab2}, our method achieved an MAE of 9.91 and an RMSE of 13.28. Compared to CLIP-Count and RCC~\cite{hobley2022learning}, our method reduced the MAE by 3.7 and over 10, respectively. Notably, our approach outperformed few-shot methods using visual prompts, highlighting its strong generalization capability.

\begin{table}
	\centering
	\setlength{\tabcolsep}{5pt}
	\footnotesize
	\begin{tabular}{clcccc}
	\toprule[1.1pt]
	\multirow{2}[2]{*}{Type} & \multirow{2}[2]{*}{Method} & \multicolumn{2}{c}{\textsc{SHB}} & \multicolumn{2}{c}{\textsc{SHA}} \\
	\cmidrule(lr){3-4} \cmidrule(lr){5-6}
	& & MAE & RMSE & MAE & RMSE \\
	\midrule
	\multirow{2}{*}{Specific} & MCNN~\cite{zhang2016single} & 85.20 & 142.30 & 221.40 & 357.80 \\
	& CrowdCLIP~\cite{liang2023crowdclip} & 69.60 & 80.70 & 217.00 & 322.70 \\
	\midrule
	\multirow{3}{*}{Generic} & RCC~\cite{hobley2022learning} & 66.60 & 104.80 & 240.10 & 366.90 \\
	& CLIP-Count~\cite{jiang2023clip} & 47.92 & 80.48 & 197.47 & 319.75 \\
	& RichCount (Ours) & \textbf{44.77} & \textbf{75.62} & \textbf{193.39} & \textbf{314.35} \\
	\bottomrule[1.1pt]
	\end{tabular}
	\caption{\textbf{Cross-Dataset Evaluation on \textsc{ShanghaiTech} Crowd Counting Dataset.} Generic models are trained on \textsc{FSC-147}, while specific models are trained on \textsc{SHA}.}
	\label{tab:3}
	\vspace{-5pt}
\end{table}

\vspace{-10pt}
\paragraph{Quantitative Results on \textsc{ShanghaiTech}.}
As shown in \cref{tab:3}, in transfer experiments on the ShanghaiTech crowd counting dataset, our method showed a slight advantage. Due to the rich information and challenges posed by crowd data, this task is particularly difficult. Nevertheless, our method outperformed other CLIP-based approaches, such as CrowdCLIP and CLIP-Count, on both SHB and SHA, with particularly notable results on the sparse SHB.

\begin{table}
	\centering
	\setlength{\tabcolsep}{3pt}
	\footnotesize
	\begin{tabular}{cccccccccc}
	\toprule[1.1pt]
	\multirow{2}[2]{*}{FFN} & \multirow{2}[2]{*}{Ada} & \multirow{2}[2]{*}{Des} & \multirow{2}[2]{*}{$\mathcal{L}_c$} &
	\multicolumn{2}{c}{Val Set} & \multicolumn{2}{c}{Test Set} &
	\multicolumn{2}{c}{Average} \\
	\cmidrule(lr){5-6} \cmidrule(lr){7-8} \cmidrule(lr){9-10} 
	& & & & MAE & RMSE & MAE & RMSE & MAE & RMSE \\
	\midrule
	\Circle & \Circle & \Circle & \Circle & 19.85 & 67.69 & 17.19 & 103.44 & 18.52 & 85.57 \\
	\Circle & \CIRCLE & \Circle & \CIRCLE & 18.48 & 62.72 & {17.02} & \textbf{99.37} & 17.75 & 81.05 \\
	\Circle & \CIRCLE & \CIRCLE & \CIRCLE & \underline{17.79} & \textbf{57.13} & \underline{16.35} & 102.32 & \underline{17.07} & \underline{79.73} \\
	\midrule
	\CIRCLE & \Circle & \Circle & \Circle & 18.21 & 68.51 & 18.54 & 101.69 & 18.38 & 85.10 \\
	\CIRCLE & \CIRCLE & \Circle & \Circle & 18.13 & 61.32 & 17.57 & 104.36 & 17.85 & 82.84 \\
	\CIRCLE & \Circle & \Circle & \CIRCLE & 18.19 & {60.58} & 18.57 & 106.18 & 18.38 & 83.38 \\
	\CIRCLE & \CIRCLE & \CIRCLE & \Circle & {17.85} & 63.02 & 17.58 & 101.71 & 17.72 & 82.37 \\
	\CIRCLE & \CIRCLE & \Circle & \CIRCLE & 17.99 & 60.65 & 17.25 & {99.68} & {17.62} & {80.17} \\
	\CIRCLE & \CIRCLE & \CIRCLE & \CIRCLE & \textbf{17.68} & \underline{57.24} & \textbf{15.78} & \underline{99.65} & \textbf{16.73} & \textbf{78.45} \\
	\bottomrule[1.1pt]
	\end{tabular}
	\caption{\textbf{Ablation Study on \textsc{FSC-147}.} This study assesses the contribution of each component to the final results. Ada refers to the text adapter, Des represents ChatGPT-4, generated image descriptions, and $\mathcal{L}_c$ indicates the contrastive learning loss.}
	\label{tab:abalation_factors}
	\vspace{-5pt}
\end{table}

\subsection{Ablation Study}
\paragraph{Ablation Study on Component Contributions.}
To validate the contribution of each module in the proposed RichCount, we conducted an ablation study on the FFN, Adapter, descriptions, and contrastive loss. The results in \cref{tab:abalation_factors} show that the model incorporating all four modules achieved the best performance, underscoring the importance of each component. Excluding the FFN resulted in the second-best performance, followed by the omission of the description module. The Adapter, which aligns text and image features, and the contrastive loss were the most influential factors. Notably, incorporating enriched descriptions provided superior performance compared to using only contrastive loss. Adding the FFN led to marginal improvements, reinforcing the importance of feature alignment and enriched textual representations for boosting zero-shot performance.


\begin{table}
	\centering
	\setlength{\tabcolsep}{7pt}
	\footnotesize
	\begin{tabular}{lccccc}
	\toprule[1.1pt]
	\multirow{2}[2]{*}{Method} & \multirow{2}[2]{*}{Exemplar} & \multicolumn{2}{c}{Val Set} & \multicolumn{2}{c}{Test Set} \\ 
	\cmidrule(lr){3-4} 
	\cmidrule(lr){5-6}
	& & MAE & RMSE & MAE & RMSE \\
	\midrule
 Text & Text & \underline{17.99} & 60.65 & 17.25 & 99.68 \\
	Claude~\cite{TheC3} & Text & 18.42 & 62.59 & 17.13 & 102.20 \\
	GPT-4o~\cite{achiam2023gpt} & Text & {18.27} & 62.13 & 17.59 & 100.36 \\
	GPT-4 & Text & 18.65 & \underline{58.55} & \underline{16.37} & 102.48 \\
	\midrule
	Claude~\cite{TheC3} & Claude & 18.05 & {59.36} & \underline{}16.63 & 100.65 \\
	GPT-4o~\cite{achiam2023gpt} & GPT-4o & 18.08 & 65.29 & 16.85 & \textbf{98.84} \\
	GPT-4~\cite{achiam2023gpt} & GPT-4 & \textbf{17.68} & \textbf{57.24} & \textbf{15.78} & \underline{99.65} \\
	\bottomrule[1.1pt]
	\end{tabular}
	\caption{\textbf{Impact of Image Descriptions Generated by GPT-4, GPT-4-turbo, and Claude on Counting Performance on \textsc{FSC-147}.}}
	\label{tab:abalation_de}
	\vspace{-5pt}
\end{table}

\vspace{-10pt}
\paragraph{Ablation Study on MLLMs.}
Expanded descriptions play a critical role in enhancing the counting model’s ability to handle flexible text-based object counting. \cref{tab:abalation_de} presents the results of experiments using image descriptions generated by ChatGPT-4~\cite{achiam2023gpt}, ChatGPT-4-turbo~\cite{achiam2023gpt}, and Claude~\cite{TheC3}. ChatGPT-4 achieved the best performance overall. Compared to directly using the original \textsc{FSC-147} category text, both Claude and ChatGPT-4-turbo slightly increased the error on the Val Set but reduced the error on the test set, demonstrating the effectiveness of expanded descriptions for unseen categories. Notably, descriptions generated by ChatGPT-4 significantly improved counting performance, emphasizing the value of rich, descriptive information.

\begin{figure}
	\centering
	\includegraphics[width = \linewidth]{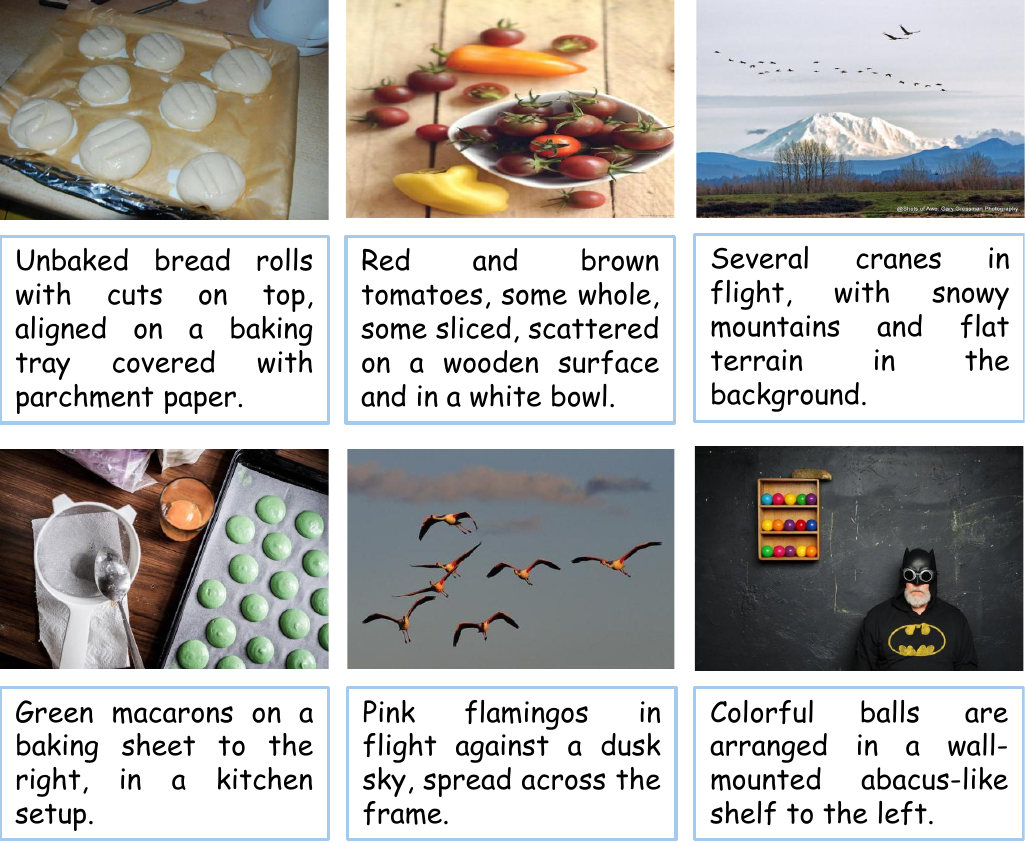}
	\caption{\textbf{Illustration of Descriptions Generated by ChatGPT-4.}}
	\label{fig:4}
	\vspace{-5pt}
\end{figure}

\subsection{Qualitative Results}
\paragraph{Analysis of Expanded Descriptions.}
\cref{fig:4} illustrates enriched descriptions for specified categories, incorporating details such as color, shape, state, and position. For example, descriptions like ``unbaked bread rolls'' and various types of ``tomatoes'' are accurately supplemented with relevant attributes. While these descriptions rarely include explicit quantity information, they closely align with the specified categories. In the bottom-right image, for instance, the description emphasizes ``colorful balls'', omitting more prominent elements such as people, thereby enhancing the semantic depth of the simple category term.

\begin{figure}
	\centering
	\includegraphics[width = \linewidth]{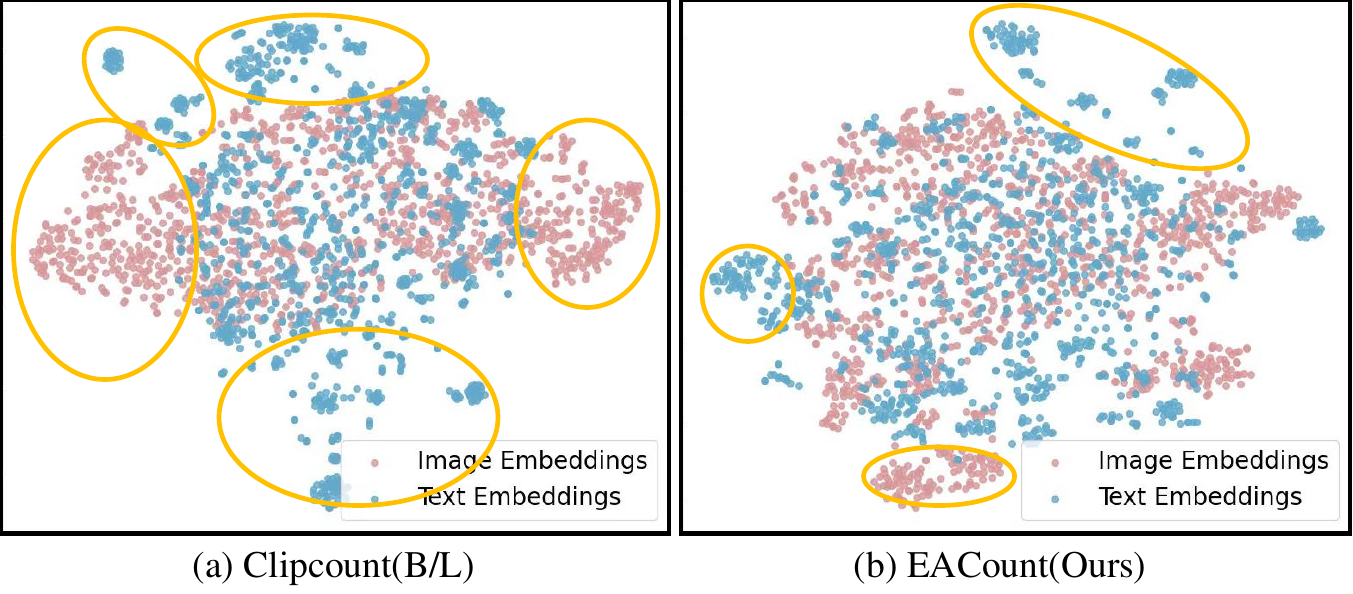}
	\caption{\textbf{Visualization of t-SNE Clusters Before and After Alignment.} Misaligned clusters are circled in yellow.}
	\label{fig:5}
	\vspace{-5pt}
\end{figure}

\vspace{-10pt}
\paragraph{Analysis of Image-Text Alignment.}
\cref{fig:5} illustrates the clustering of image-text features before and after feature alignment. In \cref{fig:5}(a), while some overlap between the features is observed, many samples remain scattered outside their respective clusters, with several image-text pairs failing to establish a correspondence. This suggests that, even when the object described by the text prompt is present in the image, the model struggles to associate them. In contrast, \cref{fig:5}(b) shows a significant reduction in misalignments, with most samples forming cohesive clusters and only a few remaining unaligned, highlighting the effectiveness of the feature alignment process.

\begin{figure*}
	\centering
	\includegraphics[width = 0.98 \linewidth]{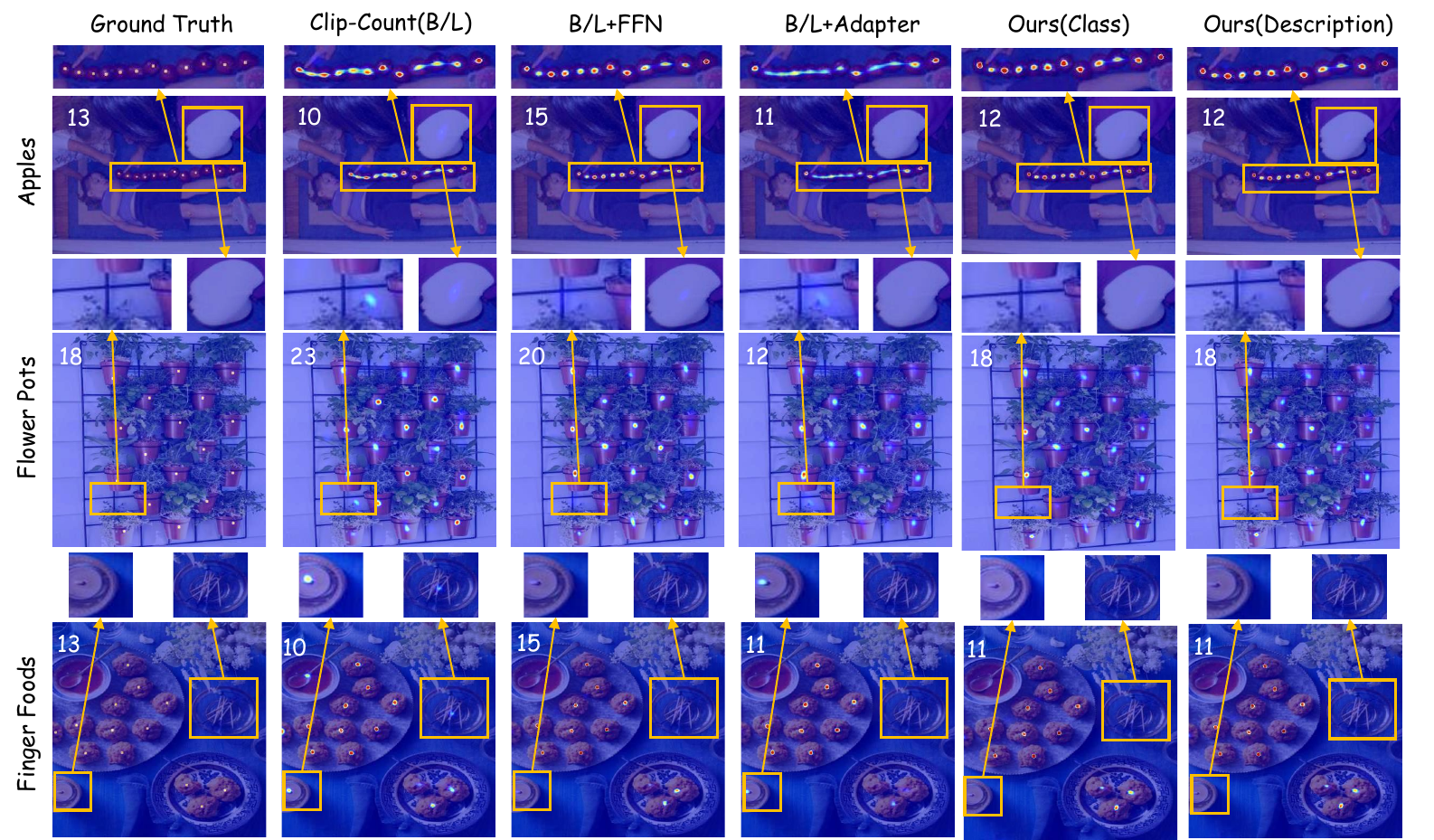}
 	\caption{\textbf{Zero-Shot Density Maps on \textsc{FSC-147}.} Errors are highlighted in orange. {B/L+FFN}: FFN in the image encoder; {B/L+Adapter}: adapter in the text encoder. {Class}: tested with class labels; {Description}: tested with image descriptions.}
	\label{fig:density1}
	\vspace{-5pt}
\end{figure*}

\vspace{-10pt}
\paragraph{Analysis of Density Map.}
\cref{fig:density1} provides a comprehensive analysis of density maps generated under various settings, demonstrating the ability of our method to reduce errors in multi-class scenarios with previously unseen categories. In the first row, our approach shows significant improvements in distinguishing dense, small objects, such as apples, from other categories, with all model variations outperforming the baseline. The second row highlights our method's ability to locate objects based on spatial cues derived from textual descriptions, addressing challenges posed by insufficiently rich text information, which prior studies have identified as a limitation for accurate counting. The third row demonstrates the model's ability to accurately count objects specified by text, such as ``Finger Foods'' or ``yellow finger foods on the plate'', even when multiple instances of the same category are present. It also shows robustness in identifying objects with distinct visual characteristics, such as sharp edges. 

\begin{figure}
	\centering
	\includegraphics[width = \linewidth]{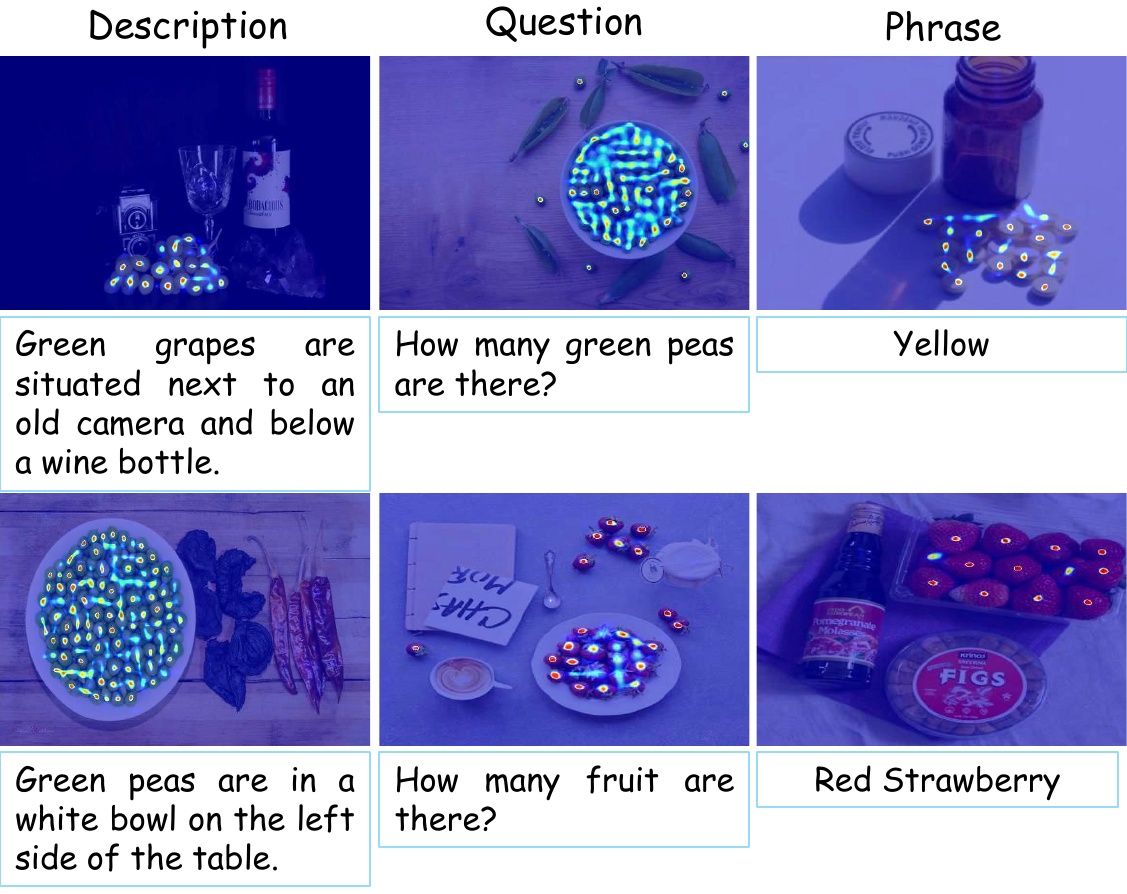}
	\caption{\textbf{Illustration of Density Maps Generated from Various Texts.}}
	\label{fig:d}
	\vspace{-5pt}
\end{figure}

\vspace{-10pt}
\paragraph{Analysis of Density Maps for Various Text Inputs.} \cref{fig:d} presents density maps generated from various text inputs, including descriptions, questions, attributes, and categories. The model consistently produces density maps aligned with the prompts, showcasing its adaptability to text-prompt-based counting. Remarkably, it can count targets when prompted with attributes like color and accurately identify specific objects, such as strawberries, in response to queries about the number of fruits in the image.

\section{Conclusion}
We propose RichCount, a two-stage framework that addresses key challenges in zero-shot object counting, including the modal gap between text and visual features and the limited semantic richness of textual prompts. RichCount leverages MLLMs to expand simple category labels into enriched descriptive texts, thereby enhancing semantic information. The framework then aligns visual and textual features using FFNs and adapters, followed by a step that enables the model to establish correspondences between text prompts and target objects in images. 
RichCount supports flexible inference, accommodating diverse text inputs such as category names, detailed descriptions, or attribute-based prompts to generate density maps. This flexibility, combined with robust feature alignment, significantly enhances the adaptability and accuracy of zero-shot counting, providing a solid foundation for future research in open-world scenarios.

\section*{Acknowledgments}
This work was supported in part by the National Natural Science Foundation of China under Grants 62472332 and 62271361, the Hubei Provincial Key Research and Development Program under Grant 2024BAB039, and the Fund of the Hubei Key Laboratory of Inland Shipping Technology under Grant NHHY2024003.


{
	\small
	\bibliographystyle{ieeenat_fullname}
	\bibliography{RichCount}

\begin{thebibliography}{35}
\providecommand{\natexlab}[1]{#1}
\providecommand{\url}[1]{\texttt{#1}}
\expandafter\ifx\csname urlstyle\endcsname\relax
  \providecommand{\doi}[1]{doi: #1}\else
  \providecommand{\doi}{doi: \begingroup \urlstyle{rm}\Url}\fi

\bibitem[Amini{-}Naieni et~al.(2023)Amini{-}Naieni, Amini{-}Naieni, Han, and Zisserman]{amini2023open}
Niki Amini{-}Naieni, Kiana Amini{-}Naieni, Tengda Han, and Andrew Zisserman.
\newblock Open-world text-specified object counting.
\newblock \emph{arXiv:2306.01851}, 2023.

\bibitem[Amini{-}Naieni et~al.(2024)Amini{-}Naieni, Han, and Zisserman]{amini2024CountGD}
Niki Amini{-}Naieni, Tengda Han, and Andrew Zisserman.
\newblock Countgd: Multi-modal open-world counting.
\newblock \emph{arXiv:2407.04619}, 2024.

\bibitem[Anthropic(2023)]{TheC3}
Anthropic.
\newblock The claude 3 model family: Opus, sonnet, haiku, 2023.

\bibitem[Arteta et~al.(2016)Arteta, Lempitsky, and Zisserman]{arteta2016counting}
Carlos Arteta, Victor~S. Lempitsky, and Andrew Zisserman.
\newblock Counting in the wild.
\newblock In \emph{Proc. Eur. Conf. Comput. Vis.}, pages 483--498, 2016.

\bibitem[Chen et~al.(2023)Chen, Zhang, Zeng, Zhang, Zhu, and Zhao]{chen2023shikra}
Keqin Chen, Zhao Zhang, Weili Zeng, Richong Zhang, Feng Zhu, and Rui Zhao.
\newblock Shikra: Unleashing multimodal llm's referential dialogue magic.
\newblock \emph{arXiv:2306.15195}, 2023.

\bibitem[Djukic et~al.(2023)Djukic, Lukezic, Zavrtanik, and Kristan]{djukic2023low}
Nikola Djukic, Alan Lukezic, Vitjan Zavrtanik, and Matej Kristan.
\newblock A low-shot object counting network with iterative prototype adaptation.
\newblock In \emph{Proc. IEEE/CVF Int. Conf. Comput. Vis.}, pages 18826--18835, 2023.

\bibitem[Hobley and Prisacariu(2022)]{hobley2022learning}
Michael~A. Hobley and Victor Prisacariu.
\newblock Learning to count anything: Reference-less class-agnostic counting with weak supervision.
\newblock \emph{arXiv:2205.10203}, 2022.

\bibitem[Hsieh et~al.(2017)Hsieh, Lin, and Hsu]{hsieh2017drone}
Meng{-}Ru Hsieh, Yen{-}Liang Lin, and Winston~H. Hsu.
\newblock Drone-based object counting by spatially regularized regional proposal network.
\newblock In \emph{Proc. {IEEE/CVF} Int. Conf. Comput. Vis.}, pages 4165--4173, 2017.

\bibitem[Huang et~al.(2024)Huang, Dai, Zhang, Zhang, and Shan]{huang2023point}
Zhizhong Huang, Mingliang Dai, Yi Zhang, Junping Zhang, and Hongming Shan.
\newblock Point, segment and count: {A} generalized framework for object counting.
\newblock In \emph{Proc. {IEEE/CVF} Conf. Comput. Vis. Pattern Recognit.}, pages 17067--17076, 2024.

\bibitem[Jiang et~al.(2023{\natexlab{a}})Jiang, Li, Ren, Liu, Zeng, Yu, and Zhang]{jiang2023t}
Qing Jiang, Feng Li, Tianhe Ren, Shilong Liu, Zhaoyang Zeng, Kent Yu, and Lei Zhang.
\newblock T-rex: Counting by visual prompting.
\newblock \emph{arXiv:2311.13596}, 2023{\natexlab{a}}.

\bibitem[Jiang et~al.(2023{\natexlab{b}})Jiang, Liu, and Chen]{jiang2023clip}
Ruixiang Jiang, Lingbo Liu, and Changwen Chen.
\newblock Clip-count: Towards text-guided zero-shot object counting.
\newblock In \emph{Proc. {ACM} Multimedia}, pages 4535--4545, 2023{\natexlab{b}}.

\bibitem[Kang et~al.(2024)Kang, Moon, Kim, and Heo]{kang2023vlcounter}
Seunggu Kang, WonJun Moon, Euiyeon Kim, and Jae{-}Pil Heo.
\newblock Vlcounter: Text-aware visual representation for zero-shot object counting.
\newblock In \emph{Proc. {AAAI} Conf. Artif. Intell.}, pages 2714--2722, 2024.

\bibitem[Lai et~al.(2024)Lai, Tian, Chen, Li, Yuan, Liu, and Jia]{lai2024lisa}
Xin Lai, Zhuotao Tian, Yukang Chen, Yanwei Li, Yuhui Yuan, Shu Liu, and Jiaya Jia.
\newblock {LISA:} reasoning segmentation via large language model.
\newblock In \emph{Proc. {IEEE/CVF} Conf. Comput. Vis. Pattern Recognit.}, pages 9579--9589, 2024.

\bibitem[Liang et~al.(2023)Liang, Xie, Zou, Ye, Xu, and Bai]{liang2023crowdclip}
Dingkang Liang, Jiahao Xie, Zhikang Zou, Xiaoqing Ye, Wei Xu, and Xiang Bai.
\newblock Crowdclip: Unsupervised crowd counting via vision-language model.
\newblock In \emph{Proc. IEEE/CVF Conf. Comput. Vis. Pattern Recognit.}, pages 2893--2903, 2023.

\bibitem[Liu et~al.(2022)Liu, Zhong, Zisserman, and Xie]{liu2022countr}
Chang Liu, Yujie Zhong, Andrew Zisserman, and Weidi Xie.
\newblock Countr: Transformer-based generalised visual counting.
\newblock In \emph{Proc. Brit. Mach. Vis. Conf.}, page 370, 2022.

\bibitem[Liu et~al.(2023)Liu, Zeng, Ren, Li, Zhang, Yang, Li, Yang, Su, Zhu, and Zhang]{Liu2023DINO}
Shilong Liu, Zhaoyang Zeng, Tianhe Ren, Feng Li, Hao Zhang, Jie Yang, Chunyuan Li, Jianwei Yang, Hang Su, Jun Zhu, and Lei Zhang.
\newblock Grounding {DINO:} marrying {DINO} with grounded pre-training for open-set object detection.
\newblock \emph{arXiv:2303.05499}, 2023.

\bibitem[Lu et~al.(2018)Lu, Xie, and Zisserman]{lu2019class}
Erika Lu, Weidi Xie, and Andrew Zisserman.
\newblock Class-agnostic counting.
\newblock In \emph{Proc. Asian Conf. Comput. Vis.}, 2018.

\bibitem[Mundhenk et~al.(2016)Mundhenk, Konjevod, Sakla, and Boakye]{mundhenk2016large}
T.~Nathan Mundhenk, Goran Konjevod, Wesam~A. Sakla, and Kofi Boakye.
\newblock A large contextual dataset for classification, detection and counting of cars with deep learning.
\newblock In \emph{Proc. Eur. Conf. Comput. Vis.}, pages 785--800, 2016.

\bibitem[OpenAI(2023)]{achiam2023gpt}
OpenAI.
\newblock {GPT-4} technical report.
\newblock \emph{arXiv:2303.08774}, 2023.

\bibitem[Paiss et~al.(2023)Paiss, Ephrat, Tov, Zada, Mosseri, Irani, and Dekel]{paiss2023teaching}
Roni Paiss, Ariel Ephrat, Omer Tov, Shiran Zada, Inbar Mosseri, Michal Irani, and Tali Dekel.
\newblock Teaching {CLIP} to count to ten.
\newblock In \emph{Proc. {IEEE/CVF} Int. Conf. Comput. Vis.}, pages 3147--3157, 2023.

\bibitem[Peng et~al.(2023)Peng, Wang, Dong, Hao, Huang, Ma, and Wei]{peng2023kosmos}
Zhiliang Peng, Wenhui Wang, Li Dong, Yaru Hao, Shaohan Huang, Shuming Ma, and Furu Wei.
\newblock Kosmos-2: Grounding multimodal large language models to the world.
\newblock \emph{arXiv:2306.14824}, 2023.

\bibitem[Ranjan et~al.(2021)Ranjan, Sharma, Nguyen, and Hoai]{ranjan2021learning}
Viresh Ranjan, Udbhav Sharma, Thu Nguyen, and Minh Hoai.
\newblock Learning to count everything.
\newblock In \emph{Proc. IEEE/CVF Conf. Comput. Vis. Pattern Recognit.}, pages 3394--3403, 2021.

\bibitem[Rasheed et~al.(2024)Rasheed, Maaz, Mullappilly, Shaker, Khan, Cholakkal, Anwer, Xing, Yang, and Khan]{rasheed2024glamm}
Hanoona~Abdul Rasheed, Muhammad Maaz, Sahal~Shaji Mullappilly, Abdelrahman~M. Shaker, Salman~H. Khan, Hisham Cholakkal, Rao~Muhammad Anwer, Eric~P. Xing, Ming{-}Hsuan Yang, and Fahad~Shahbaz Khan.
\newblock Glamm: Pixel grounding large multimodal model.
\newblock In \emph{Proc. {IEEE/CVF} Conf. Comput. Vis. Pattern Recognit.}, pages 13009--13018, 2024.

\bibitem[Ren et~al.(2024)Ren, Huang, Wei, Zhao, Fu, Feng, and Jin]{ren2024pixellm}
Zhongwei Ren, Zhicheng Huang, Yunchao Wei, Yao Zhao, Dongmei Fu, Jiashi Feng, and Xiaojie Jin.
\newblock Pixellm: Pixel reasoning with large multimodal model.
\newblock In \emph{Proc. {IEEE/CVF} Conf. Comput. Vis. Pattern Recognit.}, pages 26364--26373, 2024.

\bibitem[Sam et~al.(2022)Sam, Agarwalla, Joseph, Sindagi, Babu, and Patel]{babu2022completely}
Deepak~Babu Sam, Abhinav Agarwalla, Jimmy Joseph, Vishwanath~A. Sindagi, R.~Venkatesh Babu, and Vishal~M. Patel.
\newblock Completely self-supervised crowd counting via distribution matching.
\newblock In \emph{Proc. Eur. Conf. Comput. Vis.}, pages 186--204, 2022.

\bibitem[Shi et~al.(2022)Shi, Lu, Feng, Liu, and Cao]{shi2022represent}
Min Shi, Hao Lu, Chen Feng, Chengxin Liu, and Zhiguo Cao.
\newblock Represent, compare, and learn: {A} similarity-aware framework for class-agnostic counting.
\newblock In \emph{Proc. IEEE/CVF Conf. Comput. Vis. Pattern Recognit.}, pages 9519--9528, 2022.

\bibitem[Tyagi et~al.(2023)Tyagi, Mohapatra, Das, Makharia, Mehra, AP, and Mausam]{tyagi2023degpr}
Aayush~Kumar Tyagi, Chirag Mohapatra, Prasenjit Das, Govind Makharia, Lalita Mehra, Prathosh AP, and Mausam.
\newblock Degpr: Deep guided posterior regularization for multi-class cell detection and counting.
\newblock In \emph{Proc. IEEE/CVF Conf. Comput. Vis. Pattern Recognit.}, pages 23913--23923, 2023.

\bibitem[Wang et~al.(2023)Wang, Chen, Chen, Wu, Zhu, Zeng, Luo, Lu, Zhou, Qiao, and Dai]{wang2024visionllm}
Wenhai Wang, Zhe Chen, Xiaokang Chen, Jiannan Wu, Xizhou Zhu, Gang Zeng, Ping Luo, Tong Lu, Jie Zhou, Yu Qiao, and Jifeng Dai.
\newblock Visionllm: Large language model is also an open-ended decoder for vision-centric tasks.
\newblock In \emph{Adv. Neural Inf. Process. Syst.}, 2023.

\bibitem[Wang et~al.(2024)Wang, Xiao, Cao, and Lu]{WangX0024}
Zhicheng Wang, Liwen Xiao, Zhiguo Cao, and Hao Lu.
\newblock Vision transformer off-the-shelf: {A} surprising baseline for few-shot class-agnostic counting.
\newblock In \emph{Proc. AAAI Conf. Artif. Intell.}, pages 5832--5840, 2024.

\bibitem[Xu et~al.(2023{\natexlab{a}})Xu, Le, Nguyen, Ranjan, and Samaras]{xu2023zero}
Jingyi Xu, Hieu Le, Vu Nguyen, Viresh Ranjan, and Dimitris Samaras.
\newblock Zero-shot object counting.
\newblock In \emph{Proc. IEEE/CVF Conf. Comput. Vis. Pattern Recognit.}, pages 15548--15557, 2023{\natexlab{a}}.

\bibitem[Xu et~al.(2023{\natexlab{b}})Xu, Le, and Samaras]{xu2023zerol}
Jingyi Xu, Hieu Le, and Dimitris Samaras.
\newblock Zero-shot object counting with language-vision models.
\newblock \emph{arXiv:2309.13097}, 2023{\natexlab{b}}.

\bibitem[Yang et~al.(2021)Yang, Su, Hsu, and Chen]{yang2021class}
Shuo{-}Diao Yang, Hung{-}Ting Su, Winston~H. Hsu, and Wen{-}Chin Chen.
\newblock Class-agnostic few-shot object counting.
\newblock In \emph{Proc. IEEE/CVF Winter Conf. Appl. Comput. Vis.}, pages 869--877, 2021.

\bibitem[Zhang et~al.(2023)Zhang, Sun, Chen, Xiao, Shao, Zhang, Chen, and Luo]{zhang2023gpt4roi}
Shilong Zhang, Peize Sun, Shoufa Chen, Min Xiao, Wenqi Shao, Wenwei Zhang, Kai Chen, and Ping Luo.
\newblock Gpt4roi: Instruction tuning large language model on region-of-interest.
\newblock \emph{arXiv:2307.03601}, 2023.

\bibitem[Zhang et~al.(2016)Zhang, Zhou, Chen, Gao, and Ma]{zhang2016single}
Yingying Zhang, Desen Zhou, Siqin Chen, Shenghua Gao, and Yi Ma.
\newblock Single-image crowd counting via multi-column convolutional neural network.
\newblock In \emph{Proc. IEEE/CVF Conf. Comput. Vis. Pattern Recognit.}, pages 589--597, 2016.

\bibitem[Zhu et~al.(2024)Zhu, Yuan, Yang, Guo, Wang, Zhong, and He]{zhu2024zero}
Huilin Zhu, Jingling Yuan, Zhengwei Yang, Yu Guo, Zheng Wang, Xian Zhong, and Shengfeng He.
\newblock Zero-shot object counting with good exemplars.
\newblock In \emph{Proc. Eur. Conf. Comput. Vis.}, 2024.

\end{thebibliography}
}


\clearpage
\setcounter{page}{1}
\setcounter{section}{0}
\maketitlesupplementary

\section{Overview}


\begin{itemize}
	\item Evaluation of performance on \textsc{CountBench} (\cref{sec1})
	\item Extended visualizations of density maps (\cref{sec3})
	\item Analysis of various descriptions (\cref{sec4})
	\item Analysis of different margins (\cref{sec3})
	\item Analysis of different FFNs and adapters (\cref{sec5})
\end{itemize}

\section{Evaluation of performance on \textsc{CountBench}}
\label{sec1}
\cref{tab2} demonstrates the superior performance of the RichCount model compared to CLIP-Count on \textsc{CountBench}~\cite{paiss2023teaching}, particularly in its enhanced ability to interpret textual descriptions for counting tasks. RichCount achieves significantly lower Mean Absolute Error (MAE) and Root Mean Square Error (RMSE) values, 10.51 and 23.21, respectively, compared to CLIP-Count's 12.45 and 26.98. This substantial improvement highlights RichCount's enhanced capability to understand and process textual inputs effectively, resulting in more accurate object counting.

\begin{table}[h]
	\centering
	\footnotesize
	\setlength{\tabcolsep}{5pt}
	\begin{tabular}{lccccc}
	\toprule[1.1pt]
	\multirow{2}[2]{*}{Method} & \multirow{2}[2]{*}{Venue} &\multirow{2}[2]{*}{Exemplar} & \multicolumn{2}{c}{FSC $\to$ \textsc{C}} \\ 
	\cmidrule(lr){4-5} 
	& & & MAE & RMSE \\
	\midrule 	
	CLIP-Count~\cite{jiang2023clip} & ACM MM’23 & Text & 12.45 & 26.98 \\
	RichCount (Ours) & & Description & \textbf{10.51} & \textbf{23.21} \\	
	\bottomrule[1.1pt]
	\end{tabular}
	\caption{\textbf{Comparison of Our Method with State-of-the-Art Zero-Shot Approaches on \textsc{CountBench}.}}
	\label{tab2}
\end{table}

\begin{figure*}[h]
	\centering
	\includegraphics[width = 0.88 \linewidth]{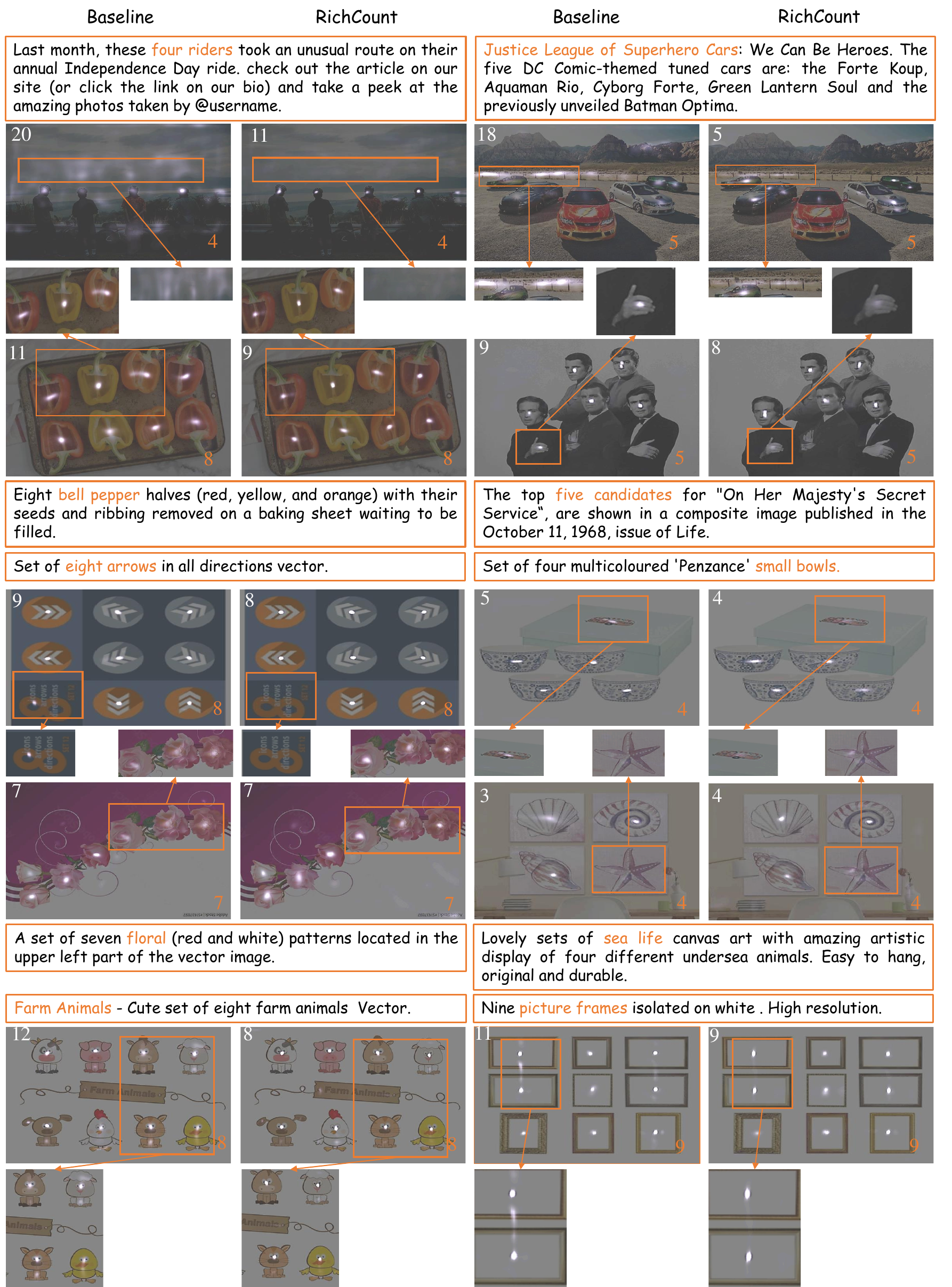} 
	\caption{\textbf{Zero-Shot Density Estimation on \textsc{CountBench} Using Models Trained on \textsc{FSC-147}.} Text inputs are sourced from \textsc{CountBench} test set, with prediction errors highlighted in orange. Predicted values are displayed in white, and ground truth values are indicated in orange.}
	\label{fig:countbench}
\end{figure*}

\cref{fig:countbench} visualizes the comparative performance of RichCount against the baseline method, highlighting our approach's superior ability to distinguish specified categories.

\section{Extended visualizations of density maps}
\label{3}
\begin{figure*}[h]
	\centering
	\includegraphics[width = 0.9 \linewidth]{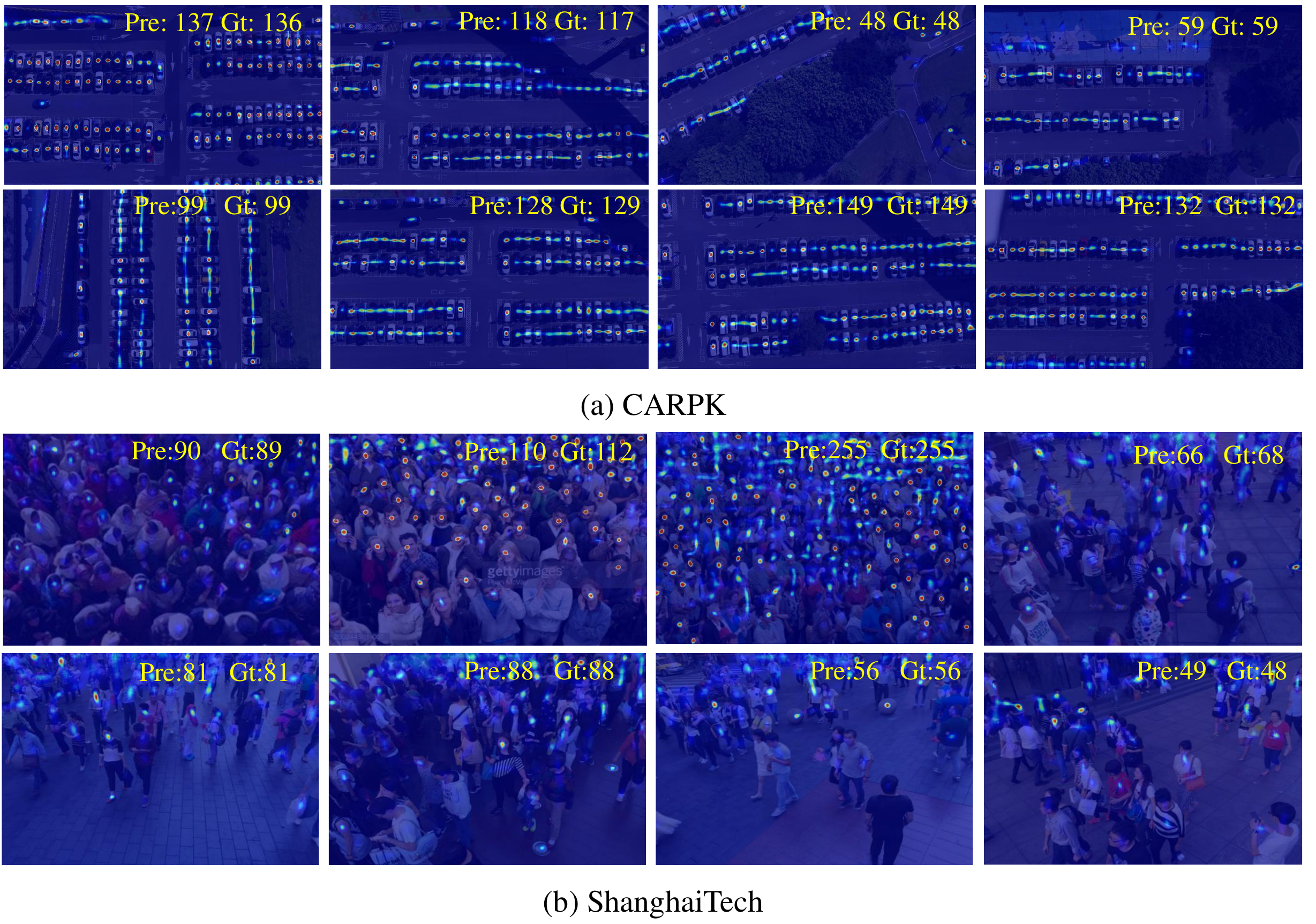}
 	\caption{\textbf{Illustration of \textsc{CARPK} and \textsc{ShanghaiTech}.}}
	\label{fig:density}
\end{figure*}

\cref{fig:density} illustrates RichCount's performance on \textsc{CARPK}~\cite{hsieh2017drone} and \textsc{ShanghaiTech}~\cite{zhang2016single}. The predicted counts (Pre) closely align with the ground truth (Gt) across parking lots on \textsc{CARPK}, demonstrating robustness in structured environments. On \textsc{ShanghaiTech}, characterized by crowded scenes, predictions remain accurate, highlighting RichCount's effectiveness in complex and dynamic crowd scenarios. 

\begin{figure*}[h]
	\centering
	\includegraphics[width = 0.9 \linewidth]{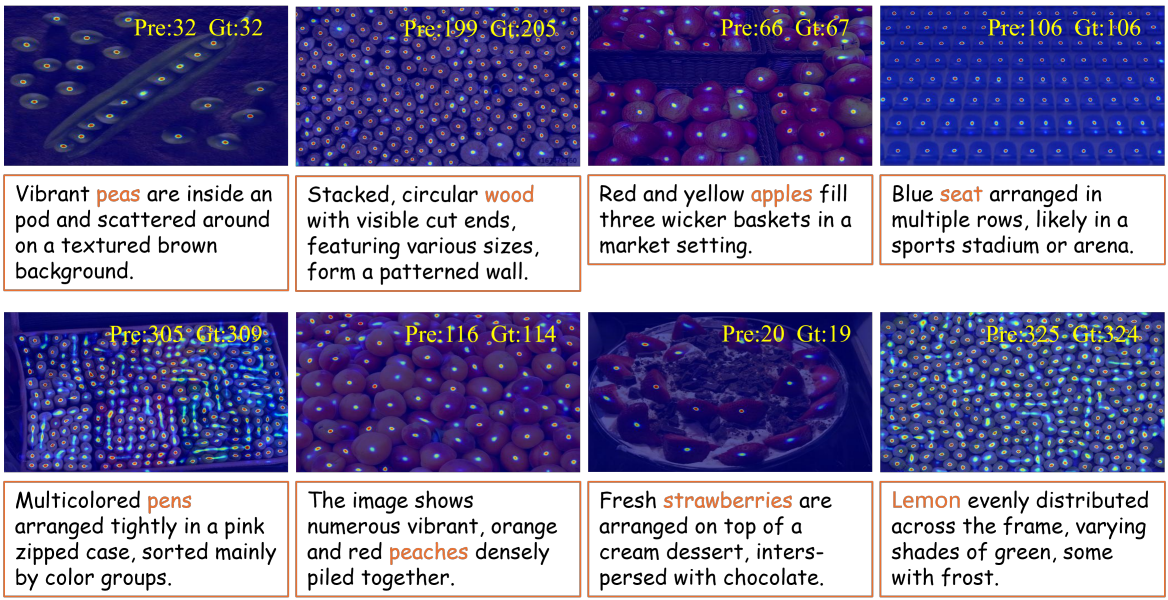}
 	\caption{\textbf{Illustration of \textsc{FSC-147}.}}
	\label{fig3}
\end{figure*}

Additionally, \cref{fig3} presents descriptions and density maps from \textsc{FSC-147}~\cite{hobley2022learning}, showcasing the model's capability to accurately count specified categories using complex textual inputs.

\section{Analysis of various descriptions}
\label{sec4}
\cref{tab:abalation_factors} presents an ablation study on \textsc{FSC-147}, comparing the use of basic category labels (\textbf{Class}), detailed descriptions (\textbf{Des}), and generic terms (\textbf{Des-f}). Models utilizing detailed descriptions consistently outperform those with simpler prompts, underscoring the importance of rich textual inputs for accurate object counting.

\begin{table}[h]
	\centering
	\setlength{\tabcolsep}{3.8pt}
	\footnotesize
	\begin{tabular}{ccccccccc}
	\toprule[1.1pt]
	\multirow{2}{*}{Class} & \multirow{2}{*}{Des} & \multirow{2}{*}{Des-f} & 
	\multicolumn{2}{c}{Val Set} & \multicolumn{2}{c}{Test Set} &
	\multicolumn{2}{c}{Average} \\
	\cmidrule(lr){4-5} \cmidrule(lr){6-7} \cmidrule(lr){8-9} 
	& & & MAE & RMSE & MAE & RMSE & MAE & RMSE \\
	\midrule
	\CIRCLE & \Circle & \Circle & 18.93 & 62.26 & 16.88 & 102.67 & 17.90 & 82.46 \\
	\CIRCLE & \CIRCLE & \Circle & \textbf{17.46} & 61.17 & 16.56 & \underline{102.19} & \underline{17.01} & 81.68\\
	\CIRCLE & \Circle & \CIRCLE & 17.88 & \underline{60.49} & \underline{16.33} & 100.42 & 17.10 & \underline{80.45} \\
	\CIRCLE & \CIRCLE & \CIRCLE & \underline{17.68} & \textbf{57.24} & \textbf{15.78} & \textbf{99.65} & \textbf{16.73} & \textbf{78.45} \\
	\bottomrule[1.1pt]
	\end{tabular}
	\caption{\textbf{Ablation Study on \textsc{FSC-147} Evaluating Various Input Texts.} {Class} uses category labels as prompts, {Des} employs descriptive sentences, and {Des-f} replaces specific category names with ``Object''.}
	\label{tab:abalation_factors}
\end{table}

\begin{figure*}[h]
	\centering
	\includegraphics[width = 0.88 \linewidth]{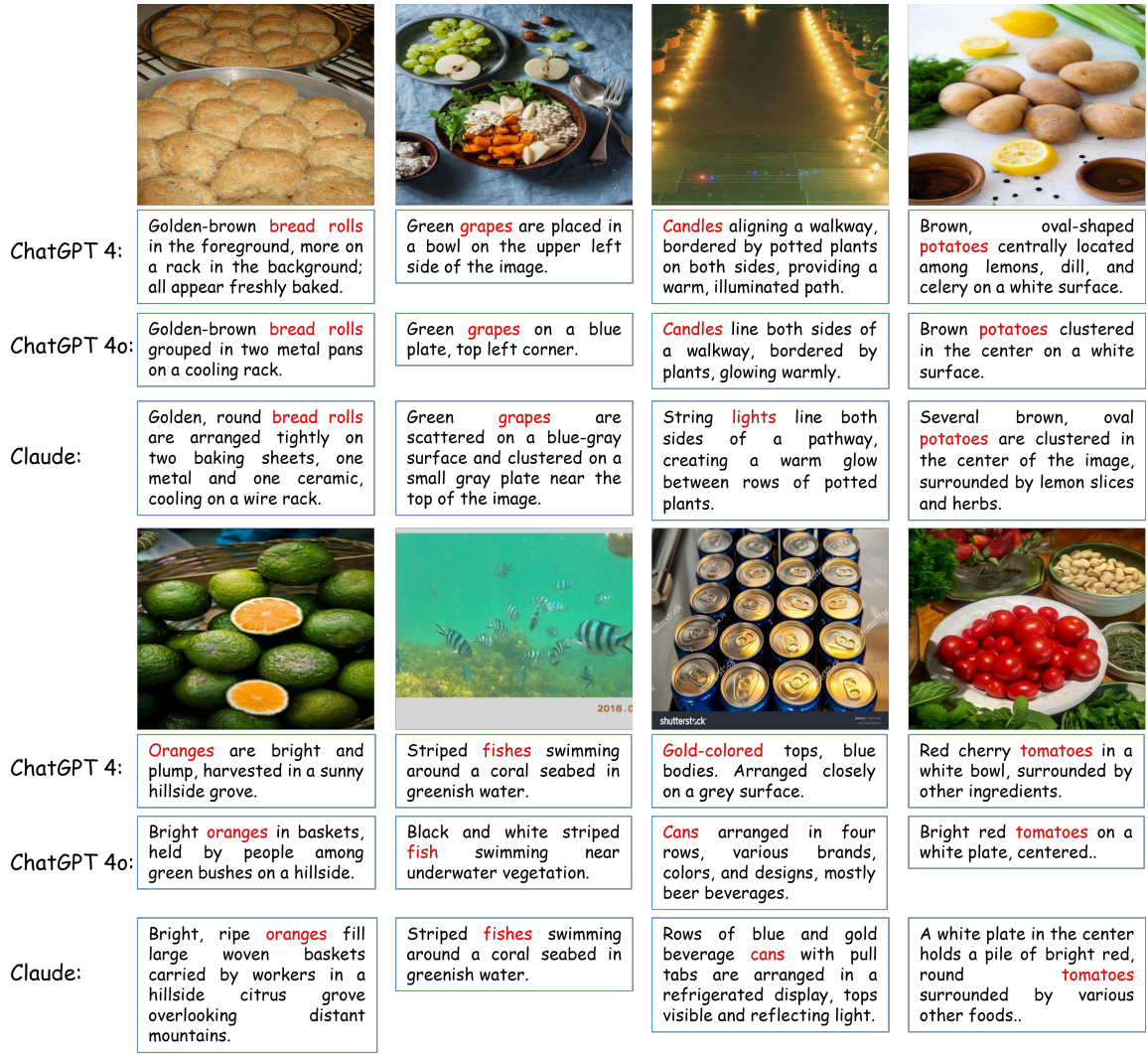}
	\caption{\textbf{Visualization of Textual Descriptions for \textsc{FSC-147} Images.} The descriptions are generated by {ChatGPT-4}, {ChatGPT-4-turbo}, and {Claude}, with count-related categories highlighted in {red}.}
	\label{fig:description}
\end{figure*}

\cref{fig:description} showcases the descriptive capabilities of large language models (ChatGPT-4~\cite{achiam2023gpt}, ChatGPT-4-turbo~\cite{achiam2023gpt}, and Claude~\cite{TheC3}) in generating text for images from \textsc{FSC-147}. While these models are generally successful in identifying the target counting categories, there are notable variations in the level of detail provided. For instance, descriptions generated by ChatGPT-4-turbo are less detailed compared to those from ChatGPT-4 and Claude. Despite these differences, the detailed and attribute-rich descriptions significantly contribute to the superior performance of RichCount. By utilizing an MSE loss that leverages these GPT-4-generated descriptions, RichCount enhances the semantic alignment between image content and textual inputs, leading to more accurate object counting.

\section{Analysis of different margins}
\label{sec3}
\cref{tab:abalation_de} illustrates the impact of various margin values on image-text alignment performance during training. We conducted a series of experiments on \textsc{FSC-147}, testing margin values of 0.2, 0.4, 0.6, 0.8, 1.0, and 1.2 over 100 epochs.

\begin{table}[h]
	\centering
	\setlength{\tabcolsep}{4pt}
	\footnotesize
	\begin{tabular}{lcccccccccc}
	\toprule[1.1pt]
	\multirow{2}{*}{Margin} & \multicolumn{2}{c}{Validation Set} & \multicolumn{2}{c}{Test Set} & \multicolumn{3}{c}{Epoch(Similarity)} \\ 
	\cmidrule(lr){2-3}
	\cmidrule(lr){4-5}
	\cmidrule(lr){6-8}
	& MAE & RMSE & MAE & RMSE & 40 & 80 & 100 \\
	\midrule
	0.2 & 18.12 & 60.78 & \underline{16.28} & \underline{101.20} & \underline{0.7663} & 0.7640 & 0.7621\\
	0.4 & 18.10 & 60.29 & 16.57 & 102.07 & \textbf{0.7686} & 0.7646 & \underline{0.7628} \\
	0.6 & 18.19 & 61.66 & 17.09 & 102.73 & 0.5687 & 0.6635 & 0.5821 \\
	0.8 & 18.14 & 63.75 & 17.43 & 101.52 & 0.3086 & 0.5211 & 0.5931 \\
	1.0 & \underline{17.68} & \underline{57.24} & \textbf{15.78} & \underline{99.65} & 0.5231 & \textbf{0.7680} & \textbf{0.7707} \\
	1.2 & \textbf{17.67} & \underline{60.58} & 16.29 & 102.25 & 0.7654 & \underline{0.7667} & 0.7267 \\
	\bottomrule[1.1pt]
	\end{tabular}
	\caption{\textbf{Effect of Varying Margin Values on Image-Text Similarity and Counting Performance on \textsc{FSC-147} Across Contrastive Training Epochs During Image-Text Alignment Experiments.}}
	\label{tab:abalation_de}
\end{table}

Using an FFN and Adapter structure within a contrastive learning framework, we observed that a margin of 0.4 effectively clustered similar image samples with their corresponding text categories while maintaining separation between different categories, achieving strong cross-modal alignment in the early stages of training. However, as training progressed, a margin of 1.0 proved more effective in bringing image and text representations closer together. This larger margin mitigates boundary ambiguity between positive and negative samples, reducing confusion among visually similar but semantically distinct categories (e.g., green grapes vs. green peas).

\section{Analysis of different FFNs and adapters}
\label{sec5}
\cref{tab:abalation_differ_ffnada} illustrates the impact of various FFN structures on the expressiveness of image and text features. Deeper or wider FFNs are capable of capturing complex feature relationships, while adapters facilitate fine-grained adjustments through variations in depth, width, or bottleneck configurations. Unlike complex FFNs, which significantly increase the number of parameters, adapters efficiently link textual prompts to semantic image information without substantial parameter expansion. By testing combinations of three-layer and five-layer FFNs and adapters, we found that a five-layer adapter paired with a five-layer FFN was the most effective in enhancing the mapping between image features and textual descriptions. This combination improves the fusion and alignment of multi-modal information, leading to more accurate object counting.

\begin{table}[h]
	\centering
	\setlength{\tabcolsep}{4pt}
	\footnotesize
	\begin{tabular}{cccccccc}
	\toprule[1.1pt]
	\multirow{2}{*}{Ada-3} & \multirow{2}{*}{Ada-5} & \multirow{2}{*}{FFN-3} & \multirow{2}{*}{FFN-5} &
	\multicolumn{2}{c}{Val Set} & \multicolumn{2}{c}{Test Set} \\
	\cmidrule(lr){5-6} \cmidrule(lr){7-8} 
	& & & & MAE & RMSE & MAE & RMSE \\
	\midrule
	 \CIRCLE & \Circle & \CIRCLE & \Circle & 17.91 & 60.28 & 16.58 & 101.16 \\
	 \Circle & \CIRCLE & \Circle & \CIRCLE & \textbf{17.68} & \textbf{57.24} & \textbf{15.78} & \textbf{99.65} \\
	\bottomrule[1.1pt]
	\end{tabular}
	\caption{\textbf{Ablation Study of FFN and Adapter Structures on \textsc{FSC-147}.} {Ada-3} employs a three-layer adapter module, whereas {Ada-5} incorporates a five-layer adapter with intermediate layers.}
	\label{tab:abalation_differ_ffnada}
\end{table}

\end{document}